\documentclass{article}
\usepackage{iclr2026_conference,times}

\usepackage[hyphens]{url}
\usepackage{graphicx}
\usepackage{amsmath}
\usepackage{amssymb}
\usepackage{forest}
\usepackage{wrapfig}
\usepackage{booktabs}
\usepackage{makecell}
\usepackage{pifont}
\usepackage[table]{xcolor}
\usepackage{algorithm}
\usepackage{algorithmic}
\usepackage{hyperref}

\definecolor{deepgreen}{RGB}{0,100,0}
\definecolor{deepred}{RGB}{150,0,0}
\newcommand{\cmark}{\ding{51}}
\newcommand{\xmark}{\ding{55}}
\newcommand{\yesmark}{\textcolor{deepgreen}{\cmark}}
\newcommand{\nomark}{\textcolor{deepred}{\xmark}}

\title{Embodied-BenchClaw: An Autonomous Multi-Agent System for Embodied Spatial Intelligence Benchmark Construction}

\author{
	\textbf{Baoyang Jiang$^{1}$ \quad
		Fengchun Zhang$^{2}$ \quad
		Leyuan Wang$^{3}$ \quad
		Haotian Li$^{3}$ \quad
		Yida Wang$^{3}$ \quad
		Zhe Ji$^{4}$} \\
	\textbf{Jinshan Lai$^{2}$ \quad
		Xi Ren$^{5}$ \quad
		Jianwei Hu$^{1}$ \quad
		Qiang Ma$^{1}$} \\
	\\
	$^{1}$QiYuan Lab \\
	$^{2}$School of Information and Software Engineering,
	\\
	 University of Electronic Science and Technology of China \\
	$^{3}$Beijing University of Posts and Telecommunications \\
	$^{4}$School of Computer Science and Engineering, Northeastern University \\
	$^{5}$School of Computer Science and Engineering, Beihang University
}

 \iclrfinalcopy

\begin{document}

\maketitle

\begin{figure}[!ht]  
	\centering
	\vspace{-4mm}
	\includegraphics[width=1\linewidth]{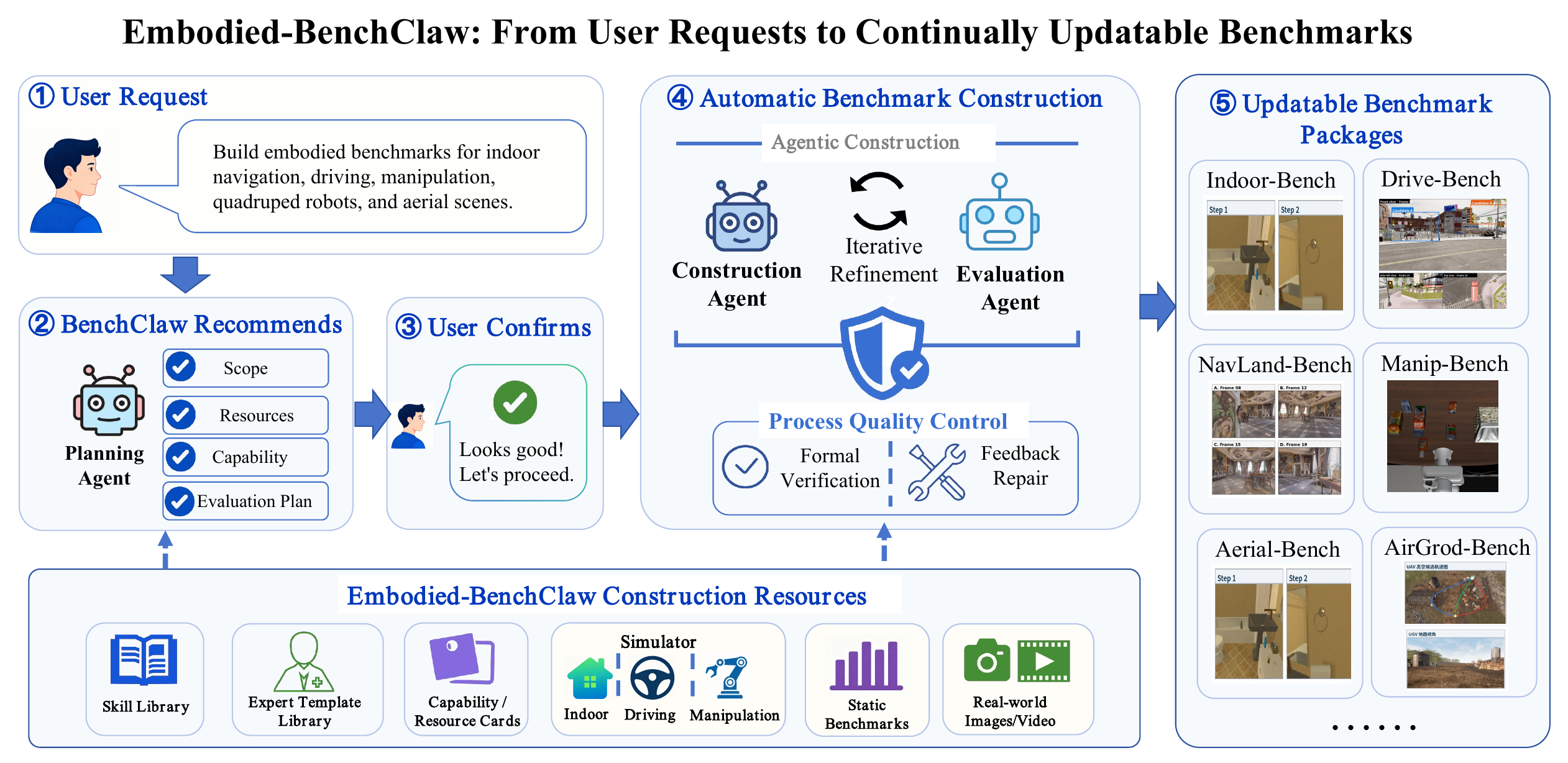}
\caption{
	Overview of Embodied-BenchClaw. 
	Given a user request, Embodied-BenchClaw recommends the benchmark scope, resources, capabilities, and evaluation plan, and then automatically constructs continually updatable benchmark packages through agentic construction and process quality control. 
	The construction process is supported by reusable resources, including skills, expert templates, capability/resource cards, simulators, static benchmarks, and real-world images or videos.
}
\label{fig:overview}
\end{figure}

\begin{abstract}
Benchmarks are essential for evaluating and advancing embodied spatial intelligence, yet their construction remains labor-intensive, hard to reuse, and difficult to maintain. Existing embodied benchmarks are often released as static datasets and may quickly become saturated as model capabilities improve, reducing their ability to distinguish next-generation models.
We propose \textbf{Embodied-BenchClaw}, an autonomous agentic system for constructing embodied spatial intelligence benchmarks. Given a user-specified evaluation intent, Embodied-BenchClaw automatically produces a complete and continually updatable benchmark package through a five-stage pipeline, including intent blueprinting, data collection, structuring and cleaning, benchmark synthesis, and evaluation reporting. The pipeline is coordinated by three agents for planning, construction, and evaluation, which respectively handle intent-to-blueprint conversion, benchmark generation, and evaluation-driven refinement. To improve reusability and reliability, Embodied-BenchClaw introduces an extensible Skill Library that decomposes complex benchmark construction into composable, verifiable, and repairable building blocks, together with process quality control to constrain and correct low-quality samples during generation.
We instantiate multiple Embodied-BenchClaw-produced benchmarks covering indoor spatial reasoning, outdoor spatial reasoning, robotic manipulation, quadruped robot navigation, UAV/aerial-view understanding, and static benchmark enhancement. These benchmarks span diverse embodied carriers, heterogeneous data sources, and fine-grained spatial capabilities. Extensive experiments, including human evaluation, judge-based quality assessment, consistency checks, cost analysis, and ablation studies, show that Embodied-BenchClaw can construct verifiable, executable, maintainable, and diagnostically useful embodied spatial benchmarks with reduced manual effort. The results demonstrate its potential for scalable and continually refreshable evaluation of embodied spatial intelligence.
\end{abstract}

\section{Introduction}
\label{sec:introduction}

Vision-language models (VLMs), multimodal large language models (MLLMs), and vision-language-action models are progressing rapidly. 
Beyond model scaling, recent research increasingly explores automated model improvement pipelines, including automatic data generation, self-evaluation, agentic feedback, and iterative capability refinement~\citep{self_rewarding_lm,self_evolving_agents,dynabench}. 
In this context, benchmarks are no longer only static leaderboards; they are key instruments for measuring emerging capabilities, exposing failure modes, and providing feedback for model evolution~\citep{dynabench}. 
However, benchmark construction remains much slower than model development. 
High-quality benchmarks are expensive to design, annotate, verify, and maintain, and once released, they may quickly become saturated, overfitted, or contaminated~\citep{dynabench,benchmark_contamination}. 
This mismatch makes it difficult to timely evaluate new model capabilities and provide reliable feedback for model improvement.

\begin{wrapfigure}{r}{0.45\linewidth}
	\centering
	\includegraphics[width=\linewidth]{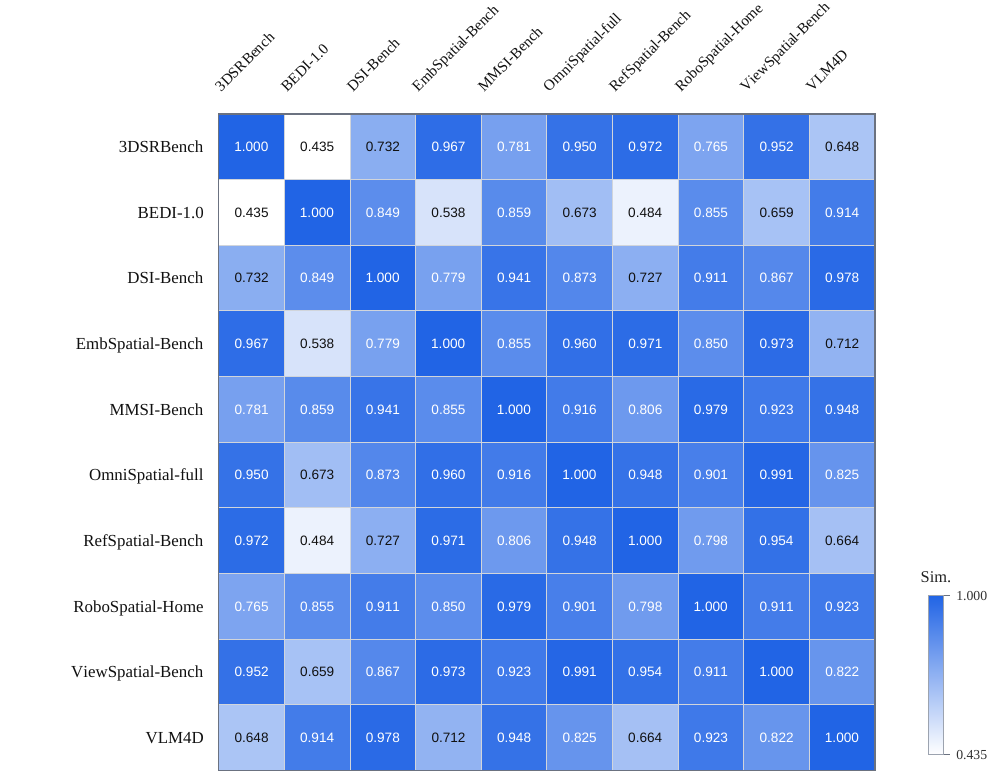}
	\caption{
		Representational similarity among embodied spatial intelligence benchmarks. 
		The similarity is computed from dataset-level Qwen-SAE activation fingerprints~\cite{qwen_scope}, revealing potential redundancy and complementarity among existing benchmarks.
	}
	\label{fig:intro1}
\end{wrapfigure}
This bottleneck is especially prominent for embodied spatial intelligence. 
Embodied agents require models to reason about egocentric spatial relations, object layouts, navigation cues, cross-view correspondences, manipulation-relevant constraints, and task feasibility from visual observations. 
A large number of embodied AI benchmarks have been developed for navigation, instruction following, manipulation, 3D understanding, autonomous driving, robotic manipulation, and multimodal agent evaluation, as summarized in recent surveys~\citep{duan2022survey,liu2025aligning}. 
Representative benchmarks such as ALFRED, LIBERO, EmbodiedScan, and EmbodiedBench provide valuable task definitions, data resources, and evaluation protocols~\citep{alfred,libero,embodiedscan,embodiedbench}. 
Many existing systems also automate parts of benchmark or data construction, such as procedural scene generation, template-based task instantiation, demonstration synthesis, trajectory replay, simulator-state verification, task generation, or reward generation~\citep{procthor,robocasa,mimicgen,eureka}. 
Nevertheless, most embodied benchmarks are still released as fixed artifacts tied to specific simulators, datasets, task definitions, or evaluation scripts. 
When new VLMs, application scenarios, or spatial capabilities emerge, constructing a suitable benchmark still requires substantial manual effort in capability decomposition, resource selection, evidence checking, item design, scoring construction, and result analysis.

Moreover, existing embodied spatial benchmarks may contain overlapping representational coverage. 
As shown in Figure~\ref{fig:intro1}, we analyze the representational similarity among existing embodied spatial intelligence benchmarks using Qwen-SAE activation fingerprints~\cite{qwen_scope}. 
For each benchmark, multimodal samples are mapped into sparse internal activation profiles, which are then averaged into dataset-level fingerprints. 
The pairwise cosine similarity heatmap shows that several benchmarks occupy highly similar regions in the model's internal feature space, suggesting potential redundancy in current embodied spatial evaluation. 
This observation motivates a construction framework that can not only generate new benchmarks, but also analyze, refine, and update benchmark packages toward under-covered capability regions.

To address this gap, we propose \textbf{Embodied-BenchClaw}, a quality-guided multi-agent framework for automatically constructing VLM-oriented embodied spatial benchmarks. 
As shown in Figure~\ref{fig:overview}, Embodied-BenchClaw transforms a high-level user request into continually updatable benchmark packages. 
It first interacts with the user to recommend and confirm the benchmark scope, resource selection, target capabilities, and evaluation plan. 
It then performs automatic benchmark construction through three cooperating agents: a planning agent that converts user intents into construction plans, a construction agent that executes the five-stage benchmark construction workflow, and a process quality-control agent that verifies intermediate artifacts and triggers local repair when validation fails. 
The construction process is supported by reusable resources, including a skill library, an expert template library, capability/resource cards, simulators, static benchmarks, and real-world images or videos.

Unlike one-shot question generation, Embodied-BenchClaw emphasizes resource-aware construction, hierarchical skill-guided execution, evidence-grounded item synthesis, executable scoring, provenance tracing, and local repair. 
Each construction stage is implemented through stage-wise skill DAGs, where skills are executable units with input-output contracts, tool bindings, validation rules, and optional repair actions. 
The process quality-control agent invokes executable verification to check intermediate artifacts, including schema validity, evidence completeness, answer derivability, scoring executability, response parsability, and trace completeness. 
When verification fails, Embodied-BenchClaw locates the affected construction step through provenance records and triggers local repair rather than restarting the full workflow. 
The final output is not merely a set of isolated questions, but a task-specific benchmark package containing items, metadata, evidence records, reference answers, scoring protocols, model response logs, evaluation reports, and update suggestions. 
In the current implementation, Embodied-BenchClaw focuses on image-, multi-view-, and simulator-state-grounded embodied spatial evaluation, rather than closed-loop robot control, tactile sensing, force feedback, or real-time physical interaction. 
By making benchmark construction reusable, verifiable, and maintainable, Embodied-BenchClaw supports timely VLM evaluation, benchmark refinement, and lifecycle-oriented benchmark maintenance.

Our contributions are summarized as follows:
\begin{itemize}
    \item We propose \textbf{Embodied-BenchClaw}, a fully automated multi-agent framework for embodied spatial benchmark construction. 
    It consists of a planning agent, a construction agent, and a process quality-control agent, enabling the transformation from user-specified evaluation intents to continually updatable benchmark packages.

    \item We design a hierarchical skill-guided execution mechanism with stage-wise DAGs. 
    Each construction stage is executed through reusable skills with explicit input-output contracts, executable components, validation rules, and repair actions, supporting modular and traceable benchmark construction.

    \item We introduce a process quality-control mechanism based on executable verification, provenance tracing, and local repair. 
    This mechanism constrains agent behavior, checks intermediate artifacts, and improves the reliability and maintainability of benchmark construction.

    \item We instantiate six representative embodied spatial benchmarks, including \textbf{Indoor-Bench}, \textbf{NavLand-Bench}, \textbf{Drive-Bench}, \textbf{Aerial-Bench}, \textbf{Quad-Bench}, and \textbf{Manip-Bench}, covering indoor reasoning, indoor navigation, autonomous driving, UAV aerial perception, quadruped robot reasoning, and robotic manipulation.
\end{itemize}

\begin{table}[t]
\centering
\caption{
Comparison with representative benchmark construction and evaluation systems. 
``/yes'' means the system explicitly supports the capability as part of its main design. 
Multi-type Data Sources refers to benchmark construction from multiple categories of data sources, such as simulators, real-world visual data, existing benchmarks, and user-provided data.
}
\label{tab:framework_comparison}
\resizebox{\textwidth}{!}{
\begin{tabular}{lccccccccc}
\toprule
\textbf{Method}
& \textbf{Modality}
& \textbf{Domain}
& \textbf{End-to-end}
& \textbf{Reusable}
& \textbf{Embodied}
& \textbf{Multi-type}
& \textbf{Evidence /}
& \textbf{Feedback}
& \textbf{Benchmark} \\
& 
& 
& \textbf{Construction}
& \textbf{Skills}
& \textbf{Spatial}
& \textbf{Data Sources}
& \textbf{Scoring}
& \textbf{/ Repair}
& \textbf{Optimization} \\
\midrule
Dynabench (2021)
& Single-modal
& NLP
& \nomark & \nomark & \nomark & \nomark & \yesmark & \yesmark & \yesmark \\

WebArena (2024)
& Multi-modal
& Agent
& \nomark & \nomark & \nomark & \nomark & \yesmark & \yesmark & \nomark \\

RoboCasa (2024)
& Multi-modal
& Embodied AI
& \nomark & \nomark & \yesmark & \nomark & \yesmark & \nomark & \nomark \\

RoboGen (2024)
& Multi-modal
& Embodied AI
& \nomark & \nomark & \yesmark & \nomark & \yesmark & \yesmark & \nomark \\

AutoBencher (2025)
& Single-modal
& LLM Evaluation
& \yesmark & \nomark & \nomark & \nomark & \yesmark & \yesmark & \yesmark \\

BenchAgents (2025)
& Single-modal
& LLM Evaluation
& \yesmark & \nomark & \nomark & \nomark & \yesmark & \yesmark & \nomark \\

EmbodiedBench (2025)
& Multi-modal
& Embodied AI
& \nomark & \nomark & \yesmark & \nomark & \yesmark & \nomark & \nomark \\

Code2Bench (2026)
& Multi-modal
& Software Engineering
& \yesmark & \nomark & \nomark & \nomark & \yesmark & \yesmark & \yesmark \\

\textbf{Embodied-BenchClaw (Ours)}
& \textbf{Multi-modal}
& \textbf{Embodied AI}
& \textbf{\yesmark} & \textbf{\yesmark} & \textbf{\yesmark} & \textbf{\yesmark} 
& \textbf{\yesmark} & \textbf{\yesmark} & \textbf{\yesmark} \\
\bottomrule
\end{tabular}
}
\end{table}

\section{Related Work}
\label{sec:related_work}

Automated benchmark construction has recently emerged as an important direction for evaluating rapidly evolving models. 
Dynabench studies human-and-model-in-the-loop dynamic benchmarking~\citep{dynabench}, while AutoBencher, BenchBench, and BenchAgents automate benchmark creation for LLM or multimodal capabilities~\citep{autobencher,benchbench,benchagents}. 
BenchAgents is particularly related to our work: it decomposes benchmark creation into planning, generation, data verification, and evaluation agents, and uses human-in-the-loop feedback to control benchmark diversity and quality~\citep{benchagents}. 
However, its main focus is general generative capabilities such as planning and constraint satisfaction during text generation, rather than embodied spatial benchmark construction.
Other systems focus on executable or environment-grounded evaluation. 
Code2Bench and PRDBench construct code-oriented benchmarks with executable tests~\citep{code2bench,prdbbench}; WebArena, SWE-bench, and Claw-Eval-Live evaluate agents in web, software, or workflow environments~\citep{webarena,swebench,clawevallive}. 
A2Eval is closer to benchmark optimization, as it reorganizes existing embodied VLM benchmarks for more compact and efficient evaluation~\citep{a2eval}. 
Table~\ref{tab:framework_comparison} summarizes the comparison.

\section{Embodied-BenchClaw}
\label{sec:benchclaw}

\subsection{Overview}
\label{sec:overview}

Embodied-BenchClaw is an automated benchmark construction framework for VLM-oriented embodied spatial intelligence. 
Given a high-level user request, Embodied-BenchClaw first interacts with the user to clarify the target capability scope, recommended data sources, evaluation settings, and model roster. 
After the user confirms the construction plan, Embodied-BenchClaw invokes reusable skills, expert templates, capability cards, and quality-feedback mechanisms to construct a complete benchmark package.

The output of Embodied-BenchClaw is not a set of isolated questions. 
Instead, it is a benchmark package containing benchmark items, metadata, evidence records, reference answers, scoring protocols, model responses, leaderboards, diagnostic reports, and update suggestions. 
This design allows the constructed benchmark to support both immediate VLM evaluation and later benchmark refinement.

\begin{figure}[t]
	\centering
	\includegraphics[width=\textwidth]{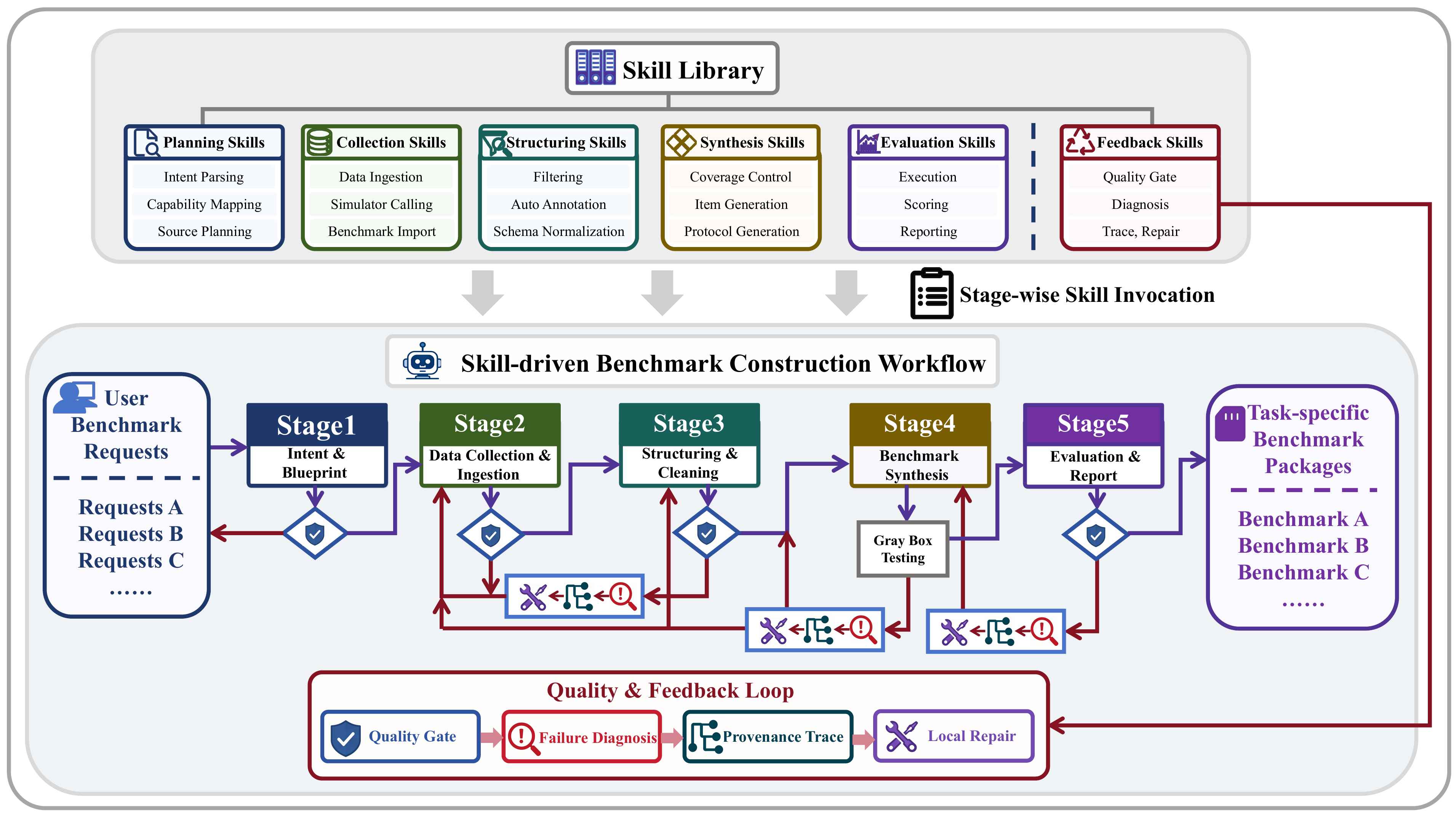}
	\caption{
		Skill-driven benchmark construction workflow with quality feedback. 
		Embodied-BenchClaw invokes stage-specific skills to transform user benchmark requests into task-specific benchmark packages through five stages: intent blueprinting, data collection, structuring and cleaning, benchmark synthesis, and evaluation reporting. 
		Quality gates, failure diagnosis, provenance tracing, and local repair enable targeted feedback to earlier stages.
	}
	\label{fig:workflow}
\end{figure}

Figure~\ref{fig:workflow} illustrates the overall workflow. 
Embodied-BenchClaw organizes benchmark construction into five automated stages, while the reusable skill library provides executable construction operations across these stages. 
Quality gates are enforced at stage boundaries to detect construction errors, trace their causes, and trigger local repair when necessary. 
The following subsections first summarize the five-stage workflow, then describe the skill library and quality-feedback mechanism in detail, followed by the expert templates and capability cards that provide reusable construction knowledge.

\subsection{Automated Construction Workflow}
\label{sec:workflow}

Embodied-BenchClaw organizes benchmark construction into five stages. 
Each stage invokes a group of reusable skills and produces intermediate artifacts for the next stage. 
This staged design makes the construction process modular, auditable, and repairable, while keeping the overall workflow fully automated after user confirmation.

\paragraph{Stage 1: Intent Blueprinting.}
This stage converts a rough benchmark request into a structured benchmark blueprint, including target capabilities, task scope, candidate resources, and preliminary evaluation requirements. 
It invokes planning-related skills to clarify the benchmark intent and prepare an executable construction plan.

\paragraph{Stage 2: Data Collection.}
This stage collects candidate data according to the blueprint, using simulators, real-world visual data, user-provided data, or existing benchmarks when available. 
It invokes acquisition-related skills to register data sources and collect raw samples with traceable source records.

\paragraph{Stage 3: Structuring and Cleaning.}
This stage converts heterogeneous raw data into cleaned and structured evidence records. 
It invokes evidence-related skills to normalize formats, prepare annotations, remove invalid samples, and organize metadata for later benchmark synthesis.

\paragraph{Stage 4: Benchmark Synthesis.}
This stage materializes benchmark items from expert templates and verified evidence. 
It invokes synthesis-related skills to bind templates with evidence, generate questions and reference answers, create scoring logic, and package valid benchmark items.

\paragraph{Stage 5: Evaluation Reporting.}
This stage evaluates target VLMs on the constructed benchmark package and generates diagnostic reports. 
It invokes evaluation-related skills to run model APIs, parse responses, compute scores, build leaderboards, and produce capability-wise and source-wise analyses.

\subsection{Hierarchical Skill-Guided Execution with Stage-wise DAGs}
\label{sec:skill_execution}

Embodied-BenchClaw follows an agentic execution paradigm, where a construction agent advances benchmark construction by invoking reusable skills within a fixed five-stage pipeline. 
The agent does not freely generate the workflow structure. 
Instead, each construction stage is implemented as a predefined stage-wise skill DAG, whose nodes are skills and whose edges represent artifact dependencies. 
This design makes the construction process controllable, reproducible, and traceable, while still allowing the agent to execute different skills according to the current artifacts, data sources, and construction requirements.

\paragraph{Skill definition.}
In Embodied-BenchClaw, a skill is not a prompt alone. 
It is a contract-defined execution unit that specifies its input artifacts, output artifacts, executable component, validation rule, and optional repair action. 
The executable component can be an LLM call, a Python script, a simulator adapter, an annotation tool, a scoring function, a data cleaning operator, or a validator. 
Thus, Python code is treated as part of a skill when it implements a construction, verification, or repair operation under a declared input-output contract. 
This definition separates skill execution from free-form prompting: prompts are mainly used for intent understanding, language realization, and report summarization, while evidence extraction, answer derivation, scoring, and formal checks are handled by tools, programs, and validators.

Formally, each skill is represented as
\[
\mathcal{S}_k=
\langle 
\mathcal{I}_k,
\mathcal{O}_k,
\mathcal{E}_k,
\mathcal{V}_k,
\mathcal{R}_k,
\tau_k
\rangle,
\]
where $\mathcal{I}_k$ and $\mathcal{O}_k$ denote the input and output artifact schemas, $\mathcal{E}_k$ denotes the executable component, $\mathcal{V}_k$ denotes the validation rule, $\mathcal{R}_k$ denotes the optional repair action, and $\tau_k$ denotes stage, capability, evidence, or tool tags used for skill selection. 
A skill can be executed only when its required input artifacts are available and its preconditions are satisfied.

\paragraph{Hierarchical skill roles.}
To support the full benchmark construction process, Embodied-BenchClaw groups skills into four roles. 
\emph{Control skills} manage stage transitions, artifact dependency tracking, DAG execution, and provenance logging. 
\emph{Construction skills} produce benchmark artifacts, including collected data, structured evidence, benchmark items, reference answers, scoring scripts, and evaluation reports. 
\emph{Verification skills} implement process quality control through executable checks, such as schema validation, evidence completeness checking, answer derivability testing, ambiguity checking, duplication detection, and scoring executability testing. 
\emph{Repair skills} are triggered when verification fails; they localize the responsible upstream node, revise invalid artifacts, and rerun only the affected subgraph. 
This role-based hierarchy makes quality control and feedback repair part of skill-guided execution rather than an external post-processing step.

\paragraph{Stage-wise skill DAGs.}
For each construction stage $s$, Embodied-BenchClaw uses a predefined skill DAG
\[
\mathcal{G}_s=(\mathcal{N}_s,\mathcal{A}_s),
\]
where $\mathcal{N}_s$ denotes the skill nodes in stage $s$ and $\mathcal{A}_s$ denotes artifact-dependency edges. 
A node consumes input artifacts, executes its corresponding skill, and produces output artifacts for downstream nodes. 
Independent nodes can be executed in parallel, while dependent nodes are scheduled according to the DAG order. 
Because the DAG is predefined by developers rather than generated freely by the LLM, Embodied-BenchClaw avoids unstable workflow planning while still allowing different skills to be activated according to the benchmark request and available resources.

Verification and repair are also represented as skill nodes in the stage-wise DAG. 
After key construction nodes produce artifacts, verification nodes check whether these artifacts satisfy schema constraints, evidence requirements, answer derivation rules, scoring constraints, and traceability requirements. 
When a check fails, the construction agent uses the artifact dependency graph and provenance records to locate the responsible node, invokes the corresponding repair skill, and reruns the affected subgraph. 
This local repair strategy avoids restarting the full five-stage workflow when only a local artifact is invalid.

Table~\ref{tab:skill_guided_execution} summarizes how hierarchical skill-guided execution is instantiated across the five construction stages. 
The table highlights the predefined DAG nodes in each stage, the corresponding skill roles, the artifact contracts, and the verification or repair focus.

\begin{table}[t]
\centering
\caption{
Hierarchical skill-guided execution in Embodied-BenchClaw. 
Each construction stage is implemented as a predefined skill DAG consisting of control, construction, verification, and repair skills. 
Quality control is embedded through verification and repair nodes rather than applied only as final post-processing.
}
\label{tab:skill_guided_execution}
\resizebox{\textwidth}{!}{
\begin{tabular}{p{0.12\textwidth} p{0.25\textwidth} p{0.17\textwidth} p{0.24\textwidth} p{0.22\textwidth}}
\toprule
\textbf{Stage / Loop}
& \textbf{Representative DAG Nodes}
& \textbf{Skill Roles}
& \textbf{Artifact Contract}
& \textbf{Verification / Repair Focus} \\
\midrule

\rowcolor[HTML]{F6F9FE}
Stage 1: Intent Blueprinting
& Intent parsing; capability decomposition; resource recommendation; metric drafting; workflow planning; blueprint validation.
& Control; Construction; Verification
& User request $\rightarrow$ intent blueprint, capability scope, resource plan, metric draft, execution plan.
& Checks intent consistency, capability--resource feasibility, metric alignment, and unresolved ambiguities. \\

\rowcolor[HTML]{F5FAF5}
Stage 2: Data Collection
& Source registration; simulator acquisition; real-data acquisition; user-data ingestion; existing-benchmark ingestion; source validation.
& Construction; Verification; Repair
& Resource plan $\rightarrow$ raw samples, source records, preliminary metadata, collection log.
& Checks source availability, schema validity, collection completeness, and source traceability; repairs missing or invalid source records. \\

\rowcolor[HTML]{F6FAFC}
Stage 3: Structuring and Cleaning
& Evidence normalization; annotation preparation; simulator-GT conversion; tool invocation; data cleaning; evidence validation.
& Construction; Verification; Repair
& Raw samples $\rightarrow$ unified evidence pool, annotations, cleaned records, metadata, tool logs.
& Checks evidence completeness, GT traceability, annotation validity, visibility constraints, and cleaning validity; repairs invalid or incomplete evidence records. \\

\rowcolor[HTML]{FDF9F4}
Stage 4: Benchmark Synthesis
& Template--evidence binding; question expansion; answer-program execution; item filtering; scoring-script generation; artifact packaging.
& Construction; Verification; Repair
& Evidence pool + templates $\rightarrow$ benchmark items, reference answers, evidence links, scoring scripts, benchmark package.
& Checks evidence--template compatibility, answer derivability, scoring executability, ambiguity, duplication, and package completeness; repairs invalid bindings or regenerated items. \\

\rowcolor[HTML]{FAF9FD}
Stage 5: Evaluation Reporting
& Model evaluation; response parsing; score aggregation; leaderboard construction; capability-wise analysis; report validation.
& Construction; Verification; Repair
& Benchmark package + model outputs $\rightarrow$ parsed responses, scores, leaderboard, evaluation report.
& Checks raw-output preservation, parsing validity, score reproducibility, abnormal responses, and report completeness; repairs parsing or reporting errors. \\

\rowcolor[HTML]{FDF8FB}
Cross-stage Feedback and Repair
& Gate aggregation; provenance tracing; defect localization; affected-subgraph rerun; release validation.
& Control; Verification; Repair
& Failed artifact + execution trace $\rightarrow$ defect report, repair plan, updated artifacts, release decision.
& Localizes the responsible skill node or stage, triggers targeted repair, reruns affected subgraphs, and produces PASS / REVIEW / FAIL decisions. \\

\bottomrule
\end{tabular}
}
\end{table}

This skill-guided execution design provides the main interface for extending Embodied-BenchClaw. 
When a new simulator, dataset, annotation tool, model API, or scoring protocol is introduced, developers can add or update the corresponding skill implementation under the same input-output contract. 
Embodied-BenchClaw records skill-level execution traces, including invoked skills, consumed and generated artifacts, validation results, failure causes, repair actions, runtime, and token cost. 
These traces support debugging, auditing, efficiency analysis, and later refinement of construction skills.

\subsection{Contract-guided Process Verification and Repair}
\label{sec:quality_feedback}

Process quality control in Embodied-BenchClaw is implemented as contract-guided verification over stage-wise skill DAGs, rather than as a final post-processing step. 
Inspired by verification-driven agent workflows, Embodied-BenchClaw first converts the user request and confirmed construction plan into a structured construction contract. 
The contract specifies the target capabilities, allowed resources, required evidence fields, expected artifacts, scoring constraints, and stage transition conditions. 
It serves as the shared source of truth for the construction agent, preventing semantic drift across stages and skill calls.

Each skill node is associated with an input-output contract. 
Formally, a skill is defined as
\[
\mathcal{S}_k=
\langle
\mathcal{I}_k,\mathcal{O}_k,
\mathcal{P}_k,\mathcal{Q}_k,
\mathcal{E}_k,\mathcal{R}_k
\rangle,
\]
where $\mathcal{I}_k$ and $\mathcal{O}_k$ are the input and output artifact schemas, $\mathcal{P}_k$ and $\mathcal{Q}_k$ are the precondition and postcondition, $\mathcal{E}_k$ is the executable component, and $\mathcal{R}_k$ is the repair action. 
For each stage $s$, Embodied-BenchClaw executes a predefined skill DAG $\mathcal{G}_s=(\mathcal{N}_s,\mathcal{A}_s)$, where nodes are skill executions and edges are artifact dependencies.

For each skill node $v_k$, Embodied-BenchClaw applies a process verifier
\[
\Phi(v_k)=
\phi_{\mathrm{safe}}(v_k)
\land
\phi_{\mathrm{dag}}(v_k)
\land
\phi_{\mathrm{contract}}(o_k)
\land
\phi_{\mathrm{quality}}(o_k),
\]
where $\phi_{\mathrm{safe}}$ checks whether the agent uses only allowed skills, tools, files, and resources; $\phi_{\mathrm{dag}}$ checks whether the execution follows the predefined stage DAG and satisfies all input dependencies; $\phi_{\mathrm{contract}}$ checks schema validity and artifact consistency; and $\phi_{\mathrm{quality}}$ checks benchmark-specific requirements such as evidence completeness, answer derivability, scoring executability, ambiguity, duplication, response parsability, and report completeness. 
Only artifacts that pass these verification conditions are allowed to flow to downstream nodes.

At the stage boundary, Embodied-BenchClaw aggregates node-level verification results through a quality gate:
\[
Q_s =
\prod_{v_k \in \mathcal{N}_s}
\mathbf{1}\left[\Phi(v_k)=1\right].
\]
If $Q_s=1$, the workflow proceeds to the next stage. 
Otherwise, Embodied-BenchClaw identifies the failed node set
\[
\mathcal{F}_s=\{v_k\in\mathcal{N}_s \mid \Phi(v_k)=0\}
\]
and uses provenance traces to locate the affected subgraph
\[
\mathcal{H}_s=\mathrm{Desc}(\mathcal{F}_s;\mathcal{G}_s).
\]
Repair is then performed only on $\mathcal{H}_s$ rather than on the entire workflow. 
Typical repair actions include recollecting missing data, rerunning annotation tools, revising template--evidence bindings, regenerating invalid items, repairing scoring scripts, or rerunning response parsing. 
The repaired artifacts are re-verified before being released to downstream stages.

This mechanism addresses two requirements of agentic benchmark construction. 
First, it improves agent safety by restricting skill execution to declared resources, tools, and stage-DAG dependencies. 
Second, it constrains agent behavior toward benchmark quality by ensuring that every generated item is supported by traceable evidence, has a derivable reference answer, and can be scored by an executable protocol. 
Therefore, process quality control in Embodied-BenchClaw is not merely manual inspection or LLM self-critique, but a combination of contract checking, executable verification, provenance tracing, and local repair.
\subsection{Expert Templates and Capability Cards}
\label{sec:templates_cards}

Besides executable skills, Embodied-BenchClaw maintains two types of reusable knowledge resources: expert templates and capability cards. 
Expert templates define task patterns for embodied spatial evaluation. 
Each template specifies the target capability, question form, required evidence fields, valid conditions, answer space, and scoring constraints. 
During benchmark synthesis, templates are not directly used as final questions; they are instantiated only when the required evidence is available and passes the corresponding quality gate.

Capability cards describe the resources that Embodied-BenchClaw can use during construction, including simulators, datasets, user-provided data formats, annotation tools, cleaning operators, and model APIs. 
A capability card records what a resource can provide, how it can be invoked, what input and output formats it supports, and what constraints or failure modes should be considered. 
During intent blueprinting and data collection, Embodied-BenchClaw uses capability cards to match user requests with available resources and avoid constructing benchmark items that cannot be supported by evidence.

Together, expert templates and capability cards provide the structured knowledge required by the skill library. 
Skills execute construction operations, while templates and cards specify what should be constructed and which resources can support it. 
This separation allows Embodied-BenchClaw to extend its construction capability by adding new skills, templates, or capability cards without rewriting the full benchmark construction workflow.

\section{Embodied-BenchClaw-produced Benchmarks}
\label{sec:produced_benchmarks}

Embodied-BenchClaw (E-BenchClaw) produces standardized benchmark packages for embodied spatial intelligence evaluation. 
Each package is constructed from a user-specified evaluation intent and contains benchmark items, input observations, reference answers, evidence records, scoring protocols, metadata, provenance traces, and update suggestions. 
Unlike conventional static benchmarks that are usually tied to a single data source or task format, E-BenchClaw supports both \emph{new benchmark construction} and \emph{existing benchmark enhancement}. 
The former builds new evaluation sets from heterogeneous embodied resources, while the latter reuses images or scenes from saturated benchmarks and generates more fine-grained, evidence-dependent spatial reasoning questions.

To cover representative embodied settings, we construct six types of E-BenchClaw-produced benchmarks: indoor spatial reasoning, outdoor spatial reasoning, object manipulation, quadruped robot navigation, UAV/aerial-view understanding, and static benchmark enhancement. 
These benchmarks cover diverse data sources, embodied agents, observation forms, spatial skills, and action constraints. 
This design allows us to evaluate whether E-BenchClaw can generalize beyond a single scenario and support different embodied carriers, including egocentric agents, autonomous vehicles, robotic arms, quadruped robots, and UAVs.

\begin{table}[t]
\centering
\small
\caption{Overview of E-BenchClaw-produced benchmarks. Each benchmark package contains items, observations, evidence records, reference answers, scoring protocols, metadata, provenance traces, and update suggestions.}
\label{tab:produced_benchmarks}
\resizebox{\textwidth}{!}{
\begin{tabular}{p{2.8cm} p{3.1cm} p{3.7cm} p{3.5cm} p{4.1cm} p{3.8cm}}
\toprule
\textbf{Benchmark Type} & \textbf{Primary Embodiment} & \textbf{Typical Data Sources} & \textbf{Input Forms} & \textbf{Core Spatial Skills} & \textbf{Key Constraints} \\
\midrule

Indoor Spatial Reasoning &
Egocentric agents and indoor mobile robots &
Indoor images/videos, RGB-D data, simulated indoor scenes, and indoor QA/navigation resources &
Images, videos, depth maps, semantic maps, room layouts, and target descriptions &
Object localization, directional relations, occlusion reasoning, room topology, target reachability, and path selection &
Indoor obstacles, room connectivity, field-of-view limits, and traversable areas \\

Outdoor Spatial Reasoning &
Autonomous vehicles and ground mobile platforms &
Street-view data, road videos, inspection data, open-environment images, trajectories, and maps &
Outdoor images/videos, road regions, obstacles, target areas, trajectories, and maps &
Drivable-area recognition, obstacle reasoning, target localization, distance-scale reasoning, occlusion reasoning, and path-risk assessment &
Road boundaries, physical obstacles, safety constraints, and dynamic risks \\

Object Manipulation &
Robotic arms &
Tabletop scenes, robotic-arm simulation, real-world manipulation images/videos, and detection/segmentation results &
RGB/RGB-D images, object masks, object poses, grasp points, and task instructions &
Graspability, support and containment relations, placement feasibility, operation ordering, and affordance reasoning &
Reachable workspace, collision constraints, placement stability, and container capacity \\

Quadruped Robot Navigation &
Quadruped robots and ground inspection robots &
Rough terrains, stairs, slopes, narrow passages, simulated environments, and real robot-view data &
Egocentric images/videos, terrain labels, obstacles, trajectories, and target positions &
Terrain traversability, detour/crossing decisions, passage selection, local path planning, and stability-risk judgment &
Obstacle height, slope, stair size, passage width, and locomotion limits \\

UAV / Aerial-view Understanding &
UAVs and aerial inspection platforms &
Aerial images/videos, aerial inspection data, target-search scenes, and 3D terrain or map data &
Aerial images/videos, target boxes, region annotations, flight altitude, flight routes, and maps &
Aerial localization, region membership, global-local matching, height/distance reasoning, visibility range, and flight-path feasibility &
Flight altitude, field of view, occlusion, no-fly zones, and route constraints \\

Static Benchmark Enhancement &
The original embodiment or visual input format of the source benchmark &
Existing embodied or VLM benchmarks where current models already achieve high scores &
Original images/scenes are kept unchanged; questions, answers, evidence, and scoring rules are reconstructed &
Fine-grained spatial reasoning, multi-step relation integration, counterfactual viewpoint reasoning, occlusion reasoning, and path/action feasibility judgment &
The original data remain unchanged, while questions are made more dependent on spatial evidence and deeper reasoning \\

\bottomrule
\end{tabular}
}
\end{table}

For each benchmark type, E-BenchClaw outputs a unified benchmark package rather than isolated QA samples. 
The evidence records may include image regions, object masks, depth cues, semantic maps, trajectories, camera poses, traversable regions, action constraints, or simulator states. 
The reference answers are accompanied by derivation traces, indicating how they can be obtained from visual-spatial evidence, geometric rules, simulator metadata, or constraint checking. 
This standardized output format makes the generated benchmarks inspectable, traceable, and executable for automatic evaluation.

\section{Experiments}
\label{sec:experiments}

We evaluate E-BenchClaw from two perspectives. 
First, we evaluate it as an automated benchmark construction framework, focusing on whether it can reliably convert user requests into complete, evidence-grounded, and executable benchmark packages. 
Second, we evaluate the generated benchmarks, focusing on whether they provide meaningful signals for comparing VLMs and MLLMs on embodied spatial intelligence.

\subsection{Experimental Setup}
\label{sec:exp_setup}

\paragraph{Research questions.}
Our experiments are organized around four questions:
\textbf{RQ1} asks whether E-BenchClaw can construct benchmarks across diverse embodied scenarios and data sources.
\textbf{RQ2} asks whether the generated items are reliable, evidence-grounded, spatially consistent, and executable for scoring.
\textbf{RQ3} asks whether E-BenchClaw improves construction efficiency and whether its core components are necessary.
\textbf{RQ4} asks whether the generated benchmarks provide useful diagnostic signals for evaluating embodied spatial capabilities of VLMs/MLLMs.

\paragraph{Construction configuration.}
Unless otherwise specified, E-BenchClaw uses an qwen3.6-35b-a3b LLM backbone to coordinate the planning, construction, and quality-control agents. 
All construction runs follow the same pipeline: intent blueprinting, data collection, structuring and cleaning, benchmark synthesis, and evaluation reporting. 
The system uses a reusable Skill Library for data processing, evidence extraction, spatial verification, scoring, and repair; an Expert Template Library for embodied spatial task generation; and process quality control for executable verification and local repair.

\paragraph{Evaluation dimensions.}
We summarize the evaluation dimensions in Table~\ref{tab:exp_dimensions}. 
The experiments jointly evaluate coverage, reliability, controllability, diagnostic value, and efficiency, while avoiding reliance on a single quality signal.

\begin{table}[t]
\centering
\small
\caption{Main evaluation dimensions for E-BenchClaw.}
\label{tab:exp_dimensions}
\resizebox{\textwidth}{!}{
\begin{tabular}{p{2.8cm} p{4.0cm} p{4.2cm} p{4.6cm}}
\toprule
\textbf{Dimension} & \textbf{Evaluation Focus} & \textbf{Main Metrics} & \textbf{Purpose} \\
\midrule
Scenario and Embodiment Coverage &
Whether E-BenchClaw supports diverse scenes and embodied agents &
Number of benchmark types, number of data sources, embodiment coverage, skill coverage &
To verify that the framework is not limited to a single embodied task \\

Construction Reliability &
Whether the generated packages are complete, valid, and executable &
Construction success rate, package completeness, valid yield rate, scoring-script success rate &
To verify reliable end-to-end benchmark construction \\

Evidence-grounded Quality &
Whether items are supported by traceable visual-spatial evidence &
Evidence binding rate, evidence-grounding correctness, answer derivability, human correctness &
To verify that answers are not unsupported LLM guesses \\

Spatial and Action Consistency &
Whether spatial relations and actions satisfy geometric and physical constraints &
Spatial consistency pass rate, action feasibility pass rate, GT executability rate &
To verify that generated items respect embodied constraints \\

Process Controllability &
Whether low-quality artifacts can be detected and repaired during construction &
Quality-gate pass rates, ambiguity rate, duplicate rate, repair success rate &
To verify the effectiveness of process quality control and local repair \\

Evaluation Value &
Whether the generated benchmarks can distinguish model capabilities &
Overall accuracy, skill-wise scores, difficulty consistency, model-scale consistency &
To verify model separability and diagnostic usefulness \\

Efficiency &
Whether E-BenchClaw reduces human effort and supports fast updates &
Time per valid sample, human review time, token cost, tool calls, update time &
To verify scalability and maintainability \\
\bottomrule
\end{tabular}
}
\end{table}

\paragraph{Human evaluation and judge-based evaluation.}
We conduct human evaluation on sampled items. 
Annotators inspect the input observation, question, reference answer, evidence record, and answer derivation trace, and rate each item in terms of correctness, clarity, answerability, and relevance to the target embodied spatial skill. 
Scores are assigned on a 1--5 scale and normalized to 0--100. 
We report the human acceptance rate as the percentage of samples whose average score exceeds a predefined threshold.

We also use LLM/VLM-as-Judge as a scalable complementary evaluation. 
At the benchmark level, we report User-Intention Alignment (UIA). 
At the item level, we report Format and Schema Quality (FSQ), Question--Answer Coherence (QAC), Context--Question Correspondence (CQC), Target Skill Dependency (TSD), and Skill-Specific Challenge (SSC). 
For embodied benchmarks, we additionally report Evidence-Grounding Correctness (EGC), Spatial-Geometric Consistency (SGC), Action Feasibility Consistency (AFC), and Ground-Truth Executability (GTE).

\paragraph{Automatic quality gates.}
During construction, E-BenchClaw applies automatic quality gates to intermediate and final artifacts. 
These gates check format validity, evidence binding, GT executability, spatial consistency, action feasibility, ambiguity, redundancy, and scoring executability. 
Invalid samples are either locally repaired or discarded. 
We report the pass rate of each quality gate, the overall valid yield rate, and the repair success rate.

\paragraph{Comparison methods.}
We compare Full E-BenchClaw with four construction baselines:
\textbf{Direct LLM Generation}, which directly prompts an LLM to generate benchmark items;
\textbf{Template-only Generation}, which instantiates questions from fixed templates;
\textbf{LLM + Template Generation}, which uses an LLM to select and fill templates but removes executable verification;
and \textbf{Human-assisted Construction}, which relies on human-written or human-revised items as a high-quality reference setting.
This comparison evaluates whether E-BenchClaw improves both construction quality and efficiency over simpler alternatives.

\paragraph{Ablation studies.}
We ablate the major components of E-BenchClaw to measure their contributions. 
\textbf{w/o Automated Planning} removes explicit intent decomposition and construction planning.
\textbf{w/o Skill Library} removes reusable construction and verification skills.
\textbf{w/o Expert Template Library} replaces embodied expert templates with generic QA templates.
\textbf{w/o Formal Verification} disables geometric, physical, and GT-executability checks.
\textbf{w/o Process Quality Control} removes stage-wise quality gates.
\textbf{w/o Local Repair} detects invalid artifacts but does not repair them.
\textbf{w/o Deduplication} removes redundancy and coverage control.
We evaluate these variants using human acceptance, judge-based scores, valid yield rate, repair success rate, construction cost, and model evaluation results.

\paragraph{Model evaluation.}
To test benchmark utility, we evaluate a set of closed-source and open-source VLMs/MLLMs on the generated benchmarks. 
For multiple-choice and short-answer items, we report accuracy or exact match. 
For grounding tasks, we report grounding accuracy or IoU when annotations are available. 
For navigation and manipulation-related items, we additionally report route validity, collision-free rate, reachability accuracy, affordance accuracy, and action-feasibility accuracy when executable checks are available. 
We further report skill-wise scores to analyze model weaknesses across direction, distance, occlusion, reachability, path planning, object relations, affordance, and multi-view reasoning.

\paragraph{Efficiency analysis.}
We report time per valid sample, human review time, token cost, number of tool calls, valid yield rate, repair rate, repair success rate, and benchmark update time. 
These metrics measure whether E-BenchClaw can reduce manual effort, improve valid sample production, and support continual benchmark refresh.

\subsection{Results on UAV / Aerial-view Understanding Benchmark}
\label{sec:uav_results}

We analyze one representative E-BenchClaw-produced benchmark for UAV/aerial-view spatial understanding. 
This benchmark contains 5,000 generated questions and covers five capability dimensions: visible object/category recognition, counting with interval choices, image-plane spatial relations, visible-area and salient-scale comparison, and depth-based near-far reasoning. 
The benchmark is instantiated into five question types: single-choice selection, multi-choice selection, binary comparison, interval selection, and ranking.

\paragraph{Overall results.}
Table~\ref{tab:uav_setting_summary} reports the average performance under three settings: vision-language evaluation, blind evaluation, and random guessing. 
The average score of vision-language models reaches $65.39\%$, while the blind setting drops to $29.82\%$, which is close to the random baseline of $33.30\%$. 
This large gap indicates that the benchmark cannot be solved mainly by language priors or answer-option bias; instead, models need to rely on aerial visual-spatial evidence.

\begin{table}[t]
\centering
\small
\caption{Average performance on the UAV/aerial-view benchmark under different evaluation settings.}
\label{tab:uav_setting_summary}
\resizebox{1\linewidth}{!}{
\begin{tabular}{lcccccc}
\toprule
\textbf{Setting} & \textbf{Overall} & \textbf{Single} & \textbf{Multi} & \textbf{Binary} & \textbf{Interval} & \textbf{Ranking} \\
\midrule
Vision-language Avg. & 65.39 & 74.79 & 52.29 & 84.36 & 32.07 & 83.46 \\
Blind Avg. & 29.82 & 30.00 & 18.00 & 43.43 & 17.21 & 40.45 \\
Random & 33.30 & 28.00 & 8.50 & 56.00 & 28.00 & 46.00 \\
\midrule
VL - Blind Gap & +35.57 & +44.79 & +34.29 & +40.93 & +14.86 & +43.01 \\
\bottomrule
\end{tabular}
}
\end{table}

\paragraph{Complete vision-language results.}
Table~\ref{tab:uav_vl_full} reports the complete results of all evaluated models under the vision-language setting. 
GPT-5.5 achieves the best overall score of $74.13\%$, followed by Gemini-3-Pro-Preview, Kimi-K2.5, Qwen3.6-27B, and Qwen3.6-35B-A3B. 
The results show that the benchmark provides meaningful model separability: strong models generally perform better, but different models exhibit different strengths across question types. 
For example, Kimi-K2.5 obtains the highest binary-comparison score, Qwen3.6-27B performs best on interval selection, and GPT-5.5 achieves the strongest overall and ranking performance.

\begin{table}[t]
\centering
\small
\caption{Complete model performance on the UAV/aerial-view benchmark under the vision-language setting. Scores are reported in percentage. Models are grouped into closed-source/API models and open-source/local models.}
\label{tab:uav_vl_full}
\resizebox{\textwidth}{!}{
\begin{tabular}{lcccccc}
\toprule
\textbf{Model} & \textbf{Overall} & \textbf{Single} & \textbf{Multi} & \textbf{Binary} & \textbf{Interval} & \textbf{Ranking} \\
\midrule
\multicolumn{7}{l}{\textit{Closed-source / API Models}} \\
\midrule
GPT-5.5 & 74.13 & 89.00 & 54.00 & 93.00 & 42.00 & 92.67 \\
Gemini-3-Pro-Preview & 71.83 & 90.00 & 54.00 & 93.00 & 33.00 & 89.17 \\
Kimi-K2.5 & 71.80 & 83.00 & 56.50 & 95.00 & 35.00 & 89.50 \\
Grok-4.20-0309-Reasoning & 68.57 & 79.00 & 54.50 & 91.00 & 33.00 & 85.33 \\
Moonshot-VL-128K-Vision-Preview & 67.53 & 82.00 & 53.00 & 84.00 & 32.00 & 86.67 \\
Claude-Opus-4-7 & 67.23 & 82.00 & 56.00 & 83.00 & 30.00 & 85.17 \\
Claude-Sonnet-4-6 & 67.07 & 79.00 & 55.00 & 85.00 & 31.00 & 85.33 \\
GPT-5.2 & 65.87 & 74.00 & 53.00 & 85.00 & 32.00 & 85.33 \\
GPT-4.1 & 63.13 & 68.00 & 63.00 & 79.00 & 24.00 & 81.67 \\
Qwen3-VL-Plus & 56.87 & 61.00 & 44.00 & 78.00 & 23.00 & 78.33 \\
Claude-Haiku-4-5-20251001 & 51.33 & 55.00 & 43.50 & 67.00 & 24.00 & 67.17 \\
Qwen3-VL-Flash-2025-10-15 & 49.87 & 50.00 & 37.50 & 67.00 & 23.00 & 71.83 \\
\midrule
\multicolumn{7}{l}{\textit{Open-source / Local Models}} \\
\midrule
Qwen3.6-27B & 71.27 & 77.00 & 55.00 & 94.00 & 44.00 & 86.33 \\
Qwen3.6-35B-A3B & 69.00 & 78.00 & 53.00 & 87.00 & 43.00 & 84.00 \\
\bottomrule
\end{tabular}
}
\end{table}

\paragraph{Complete blind and random results.}
Table~\ref{tab:uav_setting_summary} reports the complete blind-evaluation results and the random baseline. 
In the blind setting, the average score is only $29.82\%$, much lower than the vision-language average of $65.39\%$. 
This confirms that the benchmark strongly depends on visual evidence from aerial observations. 
Notably, several blind results are close to or below the random baseline, indicating that language-only guessing is insufficient for this benchmark.

\paragraph{Question-type difficulty.}
The benchmark reveals a clear difficulty hierarchy. 
Binary comparison and ranking are relatively easier, with vision-language averages of $84.36\%$ and $83.46\%$, respectively. 
Single-choice questions are also tractable, reaching $74.79\%$. 
In contrast, multi-choice questions are much harder, with an average score of $52.29\%$, because they require set-level reasoning over multiple visible targets. 
Interval selection is the most challenging type, with an average score of only $32.07\%$. 
This suggests that aerial counting, quantity-range estimation, and depth/scale-related interval reasoning remain difficult for current VLMs/MLLMs.

\paragraph{Implications.}
Overall, the UAV/aerial-view benchmark demonstrates three important properties of E-BenchClaw-produced benchmarks. 
First, the large gap between vision-language and blind evaluation confirms strong evidence dependency. 
Second, the performance differences across models show that the benchmark has meaningful discriminative power. 
Third, the large variation across question types provides fine-grained diagnostic signals, especially revealing weaknesses in multi-object selection and interval-based aerial spatial reasoning.

\bibliographystyle{iclr2026_conference}
\bibliography{references}

\end{document}